# A Robust Linguistic Platform for Efficient and Domain specific Web Content Analysis

**Thierry Hamon, Adeline Nazarenko, Thierry Poibeau, Sophie Aubin, Julien Derivière,**
LIPN – UMR CNRS 7030
99, avenue J.B. Clément
F-93430 Villetaneuse, France
*firstname.name*@lipn.univ-paris13.fr

## Abstract

Web semantic access in specific domains calls for specialized search engines with enhanced semantic querying and indexing capacities, which pertain both to information retrieval (IR) and to information extraction (IE). A rich linguistic analysis is required either to identify the relevant semantic units to index and weight them according to linguistic specific statistical distribution, or as the basis of an information extraction process. Recent developments make Natural Language Processing (NLP) techniques reliable enough to process large collections of documents and to enrich them with semantic annotations. This paper focuses on the design and the development of a text processing platform, Ogmios, which has been developed in the ALVIS project. The Ogmios platform exploits existing NLP modules and resources, which may be tuned to specific domains and produces linguistically annotated documents. We show how the three constraints of genericity, domain semantic awareness and performance can be handled all together.

## 1. Introduction

Search engines like Google or Yahoo offer access to billions of textual web pages. These tools are very popular and seem to be sufficient for a large number of general user queries on the Internet. However, some other queries are more complex, requiring specific knowledge or processing strategies: no really satisfactory solution exists for these requests. There is thus a need for more specific search engines dedicated to specialized domain or users.

Let us consider the case of text mining in Microbiology for example. Given the specificity and the reliability of the information that is sought by scientists, it is clear than one needs more than existing search engines. Even if recent developments in biology and biomedicine are reported in large bibliographical databases (e.g. Flybase, specialized on Drosophilia Menogaster or Medline), such databases and the associated searching functionalities are not sufficient to satisfy biologists' specific information needs, such as finding information on gene interactions in order to progressively figure out a whole interaction network. We previously argued that looking for this kind of relational information requires a domain-specific linguistic analysis and parsing of the documents (Alphonse *et al.*, 2004).

The ALVIS project aims at developing an open source search engine, with extended semantic search facilities. Compared to state of the art search engines (like Google, the most popular one), the ALVIS search engine is domain specific. It relies on a specialized crawler, which selects the web pages on terminological grounds. Indexing exploits various types of linguistic and domain specific annotation. Through a dedicated user interface, the ALVIS search engine processes the query more accurately, taking into account the topic and the context of search to refine both the query and the document analysis.

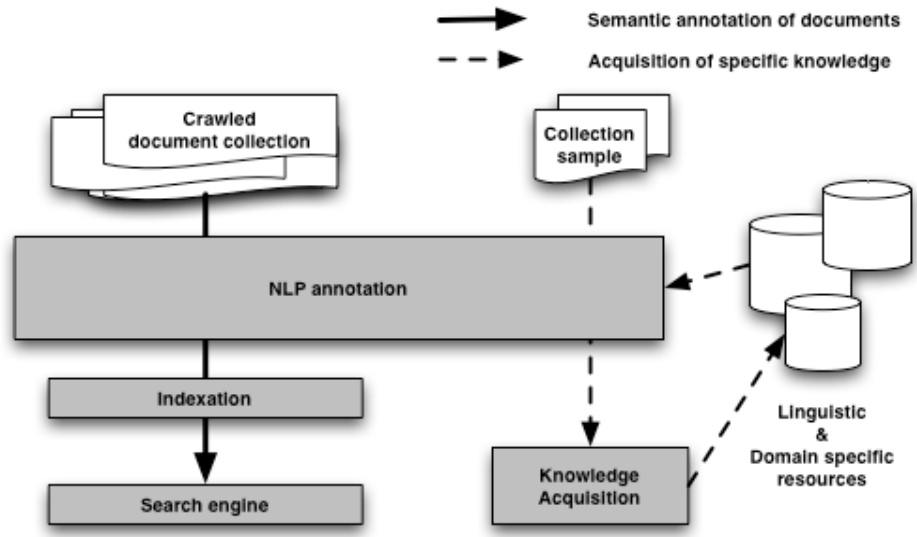

Figure 1: Role of NLP in the ALVIS semantic search engine

This paper focuses on the design and the development of the text processing platform, Ogmios, which has been developed in the ALVIS project. The challenges were
- to handle rather large domain specific collections of documents (typical specialized collections gather hundreds of thousands of documents, rather than hundreds of millions of documents),
- to analyze documents from the web using a single platform, how heterogeneous they may be,
- to enrich documents with domain-specific semantic information to allow semantic querying.

The present paper shows how the three constraints of genericity, domain semantic awareness and performance can be handled all together.

The Ogmios platform is a generic one. It is instantiated using existing NLP modules and resources, which can be tuned to specific domains. The figure 1 shows the role of the NLP annotation and resource acquisition in the whole IR process. For processing texts in the biological domain, we exploited a specific named entity dictionaries and terminologies and we adapted a generic syntactic analyzer. In this paper, we focus on parsing which is the most challenging NLP step when one wants to annotate a large document collection.

Section 2 gives an overview of the existing platforms designed for document annotation. Sections 3 and 4 describe the global architecture of the platform and its various NLP modules. Section 5 describes the performance of our system on a collection of crawled documents relative to Microbiology. The last section presents our NLP adaptation strategy, showing as an example how a specialized parser can be derived from a generic one.

## 2. Background

Several text engineering architectures have been proposed to manage text processing over the last decade (Cunningham *et al.*, 2000). GATE (General Architecture for Text Engineering) (Bontcheva *et al.*, 2004) has been essentially designed for information extraction tasks. It aims at reusing NLP tools in built-in components. The interchange annotation format (CPSL –

Common Pattern Specific Language) is based on the TIPSTER annotation format (Grishman, 1997).

Based on an external linguistic annotation platform, namely GATE, the KIM platform (Popov *et al.*, 2004) can be considered as a "meta-platform". It is designed for ontology population, semantic indexing and information retrieval. KIM has been integrated in massive semantic annotation projects such as the SWAN clusters[1] and SEKT.[2] The authors identify scalability as a critical parameter for two reasons: (1) it has to be able to process large amounts of data, in order to build and train statistical models for Information Extraction; (2) it has to support its own use as an online public service.

UIMA (Furrucci & Lally, 2004), a new implementation architecture of TEXTRACT (Neff *et al.*, 2004), is similar to GATE. It mainly differs from GATE in the data representation model. UIMA is a framework for the development of analysis engines. It offers components for the analysis of unstructured information streams as HTML web pages. These components are supposed to range from lightweight to highly scalable implementations. The UIMA SDK is a collection of Java classes. The UIMA annotation format is called CAS (Common Analysis Structure). It is mainly based on the TIPSTER format (Grishman, 1997). CAS annotations are stand-off for the sake of flexibility. Documents can be processed either at a single document level or at a collection level. Collections are handled in UIMA by the Collection Processing Engine, which has some interesting features such as filtering, performance monitoring and parallelization.

The Textpresso system (Muller *et al.*, 2004) has been specifically developed to mine biological documents, abstracts as well as articles. For instance, it has been used to process 16,000 abstracts and 3,000 full text articles related to *Caenorhabditis elegans*. It is designed as a curation system extracting gene-gene interaction that is also used as a search engine. It integrates the following NLP modules: tokenizer, sentence segmentation, Part-Of-Speech (POS) tagging, and an semantic tagging based on Gene Ontology (GOConsortium, 2001).

While Textpresso is specifically designed for biomedical texts, our platform is more similar to GATE in its aim: proposing a generic platform to process large document collections. Generally, very little information is given to evaluate the behavior of the systems on a collection of documents whereas from our point of view, this aspect is crucial for such a system. Our first test shows that GATE is not suited to process large collections of documents. GATE has been designed as a powerful environment for conception and development of NLP applications in information extraction. Scalability is not central in its design, and information extraction deals with small sets of documents. However, we have observed that problems appear on small sets of documents. We choose to propose a platform able to analyze large amounts of documents, and focus on the efficiency of the processing.

### 3. A modular and tunable platform

In the development of Ogmios, we focused on tool integration. Our initial goal was to exploit existing NLP tools rather than developing new ones[3] but integrating heterogeneous tools and nevertheless achieve good performance in document annotation was challenging. Ogmios platform was designed to test various combinations of annotations in order to identify which ones have a significant impact on information retrieval, information extraction or even extraction rule learning. In that respect, the platform can be viewed as a modular software architecture that can be configured to achieve various tasks.

---

[1] http://deri.ie/projects/swan  
[2] http://sekt.semanticweb.org  
[3] We developed NLP systems only when no other solution was available. We preferably chose GPL or free licence software when possible.

### 3.1. Specific constraints

The reuse of NLP tools imposes specific constraints regarding software engineering and processing domain-specific documents requires tuning resources to better fit the data.

From the software engineering point of view, the constraints mainly concern the input/output formats of the integrated NLP tools. Each tool has its own input and output format. Linking together several tools requires defining an interchange format. This engineering point of view is important for testing various combinations of annotations. The second type of constraints is the cost linguistic analysis in terms of processing time. The main pitfall is the deep syntactic dependency parsing which is time consuming) and which lead us to design a distributed architecture.

A domain specific annotation platform also requires lexical and ontological resources or the tuning of NLP tools such as the Part-of-Speech tagger or parser. For instance, we have argued in (Alphonse *et al.*, 2004) that identification of gene interaction requires gene name tagging, which relates to traditional named entity recognition, term recognition and a reliable syntactic analysis.

### 3.2. General architecture

The different processing steps are traditionally separated in modules (Bontcheva *et al.*, 2004). Each module carries out a specific processing step: named entity recognition, word segmentation, POS tagging, parsing, semantic tagging or anaphora resolution. It wraps an NLP tool to ensure the conformity of the input/output format with the DTD. Annotations are recorded in an XML stand-off format to deal with the heterogeneity of NLP tools input/output (the DTD is fully described in (Nazarenko *et al.*, 2006)). The modularity of the architecture simplifies the substitution of a tool by another.

Tuning to a specific field is insured by the exploitation of specialized resources by each module. For instance, a targeted species or gene list can be added to the biology-specific named entity recognizer to process Medline abstracts. In the ALVIS project, the problem of acquiring automatically these specialized resources from a training corpus is also addressed (see Figure 1 and (Alphonse *et al.*, 2004)) but this question falls out of the scope of the present paper.

Figure 2 gives an overview of the architecture. The various modules composing the NLP line are represented as boxes. The description of these modules is given in section 4. The arrows represent the data processing flow. Intermediary levels of annotations can be produced if the complete NLP line is not used. For instance, anaphora resolution is seldom activated.

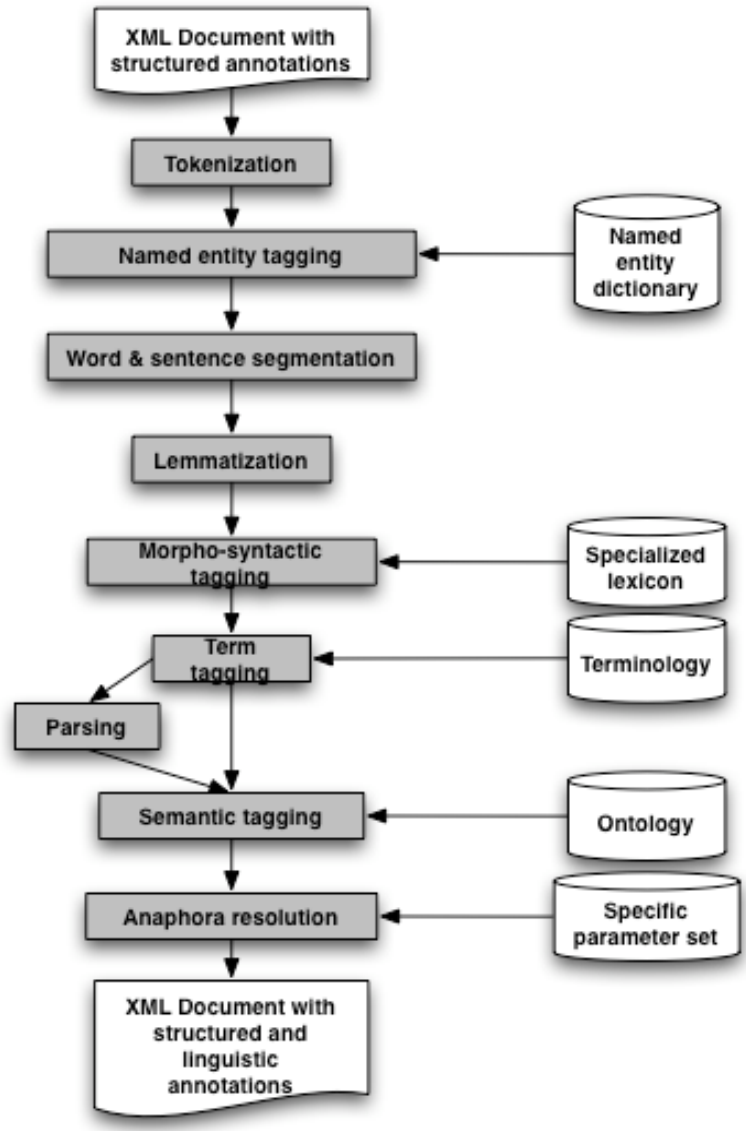

Figure 2: Ogmios architecture

We assume that input web documents are already downloaded, cleaned, encoded into the UTF-8 character set, and formatted in XML (Nazarenko *et al.*, 2006). Documents are first tokenized to define offsets to ensure the homogeneity of the various annotations. Then, documents are processed through several modules: named entity recognition, word and sentence segmentation, lemmatization, part-of-speech tagging, term tagging, parsing, semantic tagging and anaphora resolution.

Although this architecture is quite traditional, few points should be highlighted:
- Tokenization computes a first basic non-linguistic segmentation of the document, which is used for further reference. The tokens are the basic textual units in the text processing line. Tokenization serves no other purpose but to provide a starting point for segmentation. This level of annotation follows the recommendations of the TC37SC4/TEI workgroup, even if we refer to the character offset rather than pointer mark-up (TEI element ptr) in the textual signal to mark the token boundaries. To simplify further processing, we distinguish different types of tokens: alphabetical tokens, numerical tokens, separating tokens and symbolic tokens.

- Named Entity tagging takes place very early in the NLP line because unrecognized named entities hinder most NLP steps, in many sublanguages;
- Terminological tagging is used as such but is also considered as an aid for syntactic parsing. As this latter step is time consuming, we exploit the fact that terminological analysis simplifies the parsing cost.

For each document, the NLP modules are called sequentially. The outputs of the modules are stored in memory until the end of the processing. XML output is recorded at the end of the document processing.

## 4. Description of the NLP modules

This section describes the different NLP modules. Il also explains what is the expected impact of each linguistic annotation step on IR or IE performance.

*Named Entity tagging*

The Named Entity tagging module aims at annotating semantic units, with syntactic and semantic types. Each text sequence corresponding to a named entity is tagged with a unique tag corresponding to its semantic value (for example a "gene" type for gene names, "species" type for species names, etc.). We use the TagEN Named Entity tagger (Berroyer, 2004), which is based on a set of linguistic resources and grammars. Named entity tagging has a direct impact on search performance when the query contains one or two named entities, as those semantic units are have a high discriminative power.

*Word and sentence Segmentation*

This module identifies sentence and word boundaries. We use simple regular expressions, based on the algorithm proposed in (Grefenstette & Tapanainen, 1994). Part of the segmentation has been implicitly performed during the Named Entity tagging to solve some ambiguities such as the abbreviation dot in the sequence "B. subtilis", which could be understood as a full stop if it were not analyzed beforehand.

*Morpho-syntactic tagging*

This module aims at associating a part of speech (POS) tag to each word. It assumes that the word and sentence segmentation has been performed. We are using a probabilistic Part-Of-Speech tagger: TreeTagger (Schmid, 1997). The POS tags are not used as such for IR but POS tagging facilitates the rest of the linguistic processing.

*Lemmatization*

This module associates its lemma, i.e. its canonical form, to each word. The experiments presented in (Moreau, 2006) show that this morphological normalization increases the performance of search engines. If the word cannot be lemmatized (for instance a number or a foreign word), the information is omitted. This module assumes that word segmentation and morpho-syntactic information are provided. Even if it is a distinct module, we currently exploit the TreeTagger output which provides lemma as well as POS tags.

*Terminology tagging*

This module aims at recognizing the domain specific phrases in a document, like *gene expression* or *spore coat cell*. These phrases considered as the most relevant terminological items. They can be provided through terminological resources such as the Gene Ontology (GOConsortium, 2001), the MeSH (MeSH) or more widely UMLS (UMLS). They can also be acquired through corpus analysis (see Figure 1). Providing a given terminology tunes the term tagging to the corresponding domain. Previous annotation levels as lemmatization and word segmentation but also named entities are required. The goal in identifying domain specific phrases in the documents is the same as for the named entitiy recognition, *i.e.* to identify the relevant semantic units. Even if previous experiments (see (Lewis, 1992) among others) have shown a little impact of the phrases on IR performance, we argue that terminology should have a more significant impact on specialized search engines, as a terminology is relevant for a specific domain. In addition to that, a normalization procedure can associate a canonical form to any phrase occurrence (e.g. *gene expression, expression of gene, gene expressed…*). This normalization step is similar to the lemmatization one for words. Gathering associated variants under a single form modifies the phrase frequencies and thus affects IR.

*Parsing*

The parsing module aims at exhibiting the graph of the syntactic dependency relations between the words of the sentence. Parsing is a time and resource-consuming NLP, especially when compared to other NLP tasks like named entity recognition or part-of-speech tagging. As mentioned above, the syntactic analysis is especially important for the tasks that involve relations between entities (either information extraction or relational queries such as X's *speeches* as opposed to *speeches on or relative to X*). However, this technology is not yet fully compatible with Information Retrieval or Extraction.

Even if processing time is a critical point for syntactic parsing, we argue that it may enhance the semantic access to web documents. On the one hand, it is usually not necessary to parse the entire documents. A good filtering procedure may select the more relevant sections to parse. We still have to develop a method for pre-filtering the textual segments that are worth parsing as proposed in (Nédellec *et al.*, 2001). On the other hand, as we will show in Section 5, a good recognition of the terms can reduce significantly the number of possible parses and consequently the parsing processing time.

In Ogmios, the word level of annotation is required in the parser input. Depending on the choice of the parser, the morpho-syntactic level may be needed. The Link Grammar Parser (Sleator & Temperley, 1993) is integrated.

*Semantic type tagging and anaphora resolution*

The last modules are currently under test and should be integrated in the next release of the platform. The semantic type tagging associates to the previously identified semantic units tags referring to ontological concepts. This allows a semantic querying of the document base.

The anaphora resolution module establishes coreference links between the anaphoric pronoun occurrences and the antecedents they refer to. Even if solving anaphora has a small impact on the frequency counts and therefore on IE, it increases IE recall: for instance *it inhibits Y* may stand for *X inhibits Y* and must be interpreted as such in a extraction engine dealing with gene interactions.

## 5. Performance analysis

We carried out an experiment on a collection of 55,329 web documents from the biological domain. All the documents went through all NLP modules, up to the term tagging (as mentioned before, the goal is not to parse the whole documents but only some filtered part of them). A 400,000 named entity list, including species and gene names, and a 375,000 term list, issued from the MeSH and Gene Ontology have been used.

Figure 3 shows the distribution of the input document size (both axes are on a log scale). Most documents have an XML size between 1KB and 100KB. The size of the biggest document is about 5.7 MB.

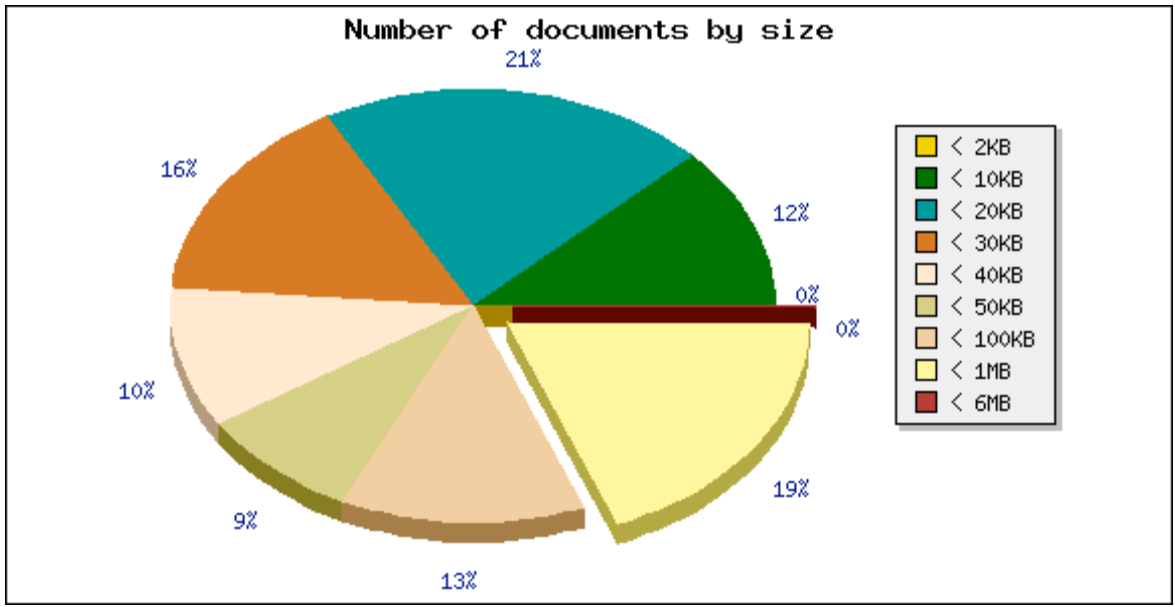

Figure 3. Distribution of document size

We used 20 machines to annotate these documents. Most of these machines were standard Personal Computers with 1GB of RAM and 2.9 or 3.1 GHz processor. We also used a computer with 8GB of RAM and two 2.8GHz Xeon (dual-core) processors. Their operating system were either Debian Linux or Mandrake Linux. The server and three NLP clients were running on the 8GB/biprocessor. Only one NLP client was running on each standard Personal Computer.

Even if a real benchmark requires several tests to evaluate the performance, we consider this performance as an interesting indication of the platform processing time. Timers are run between each function call in order to measure how long each step is (user-time-wise). We used the functions provided in the `Time::Hires` Perl package. All the time results are recorded in the annotated XML documents.

|  | Average number of units by document | Total number of units in the document collection |
|---|---|---|
| Tokens | 5,021.9 | 277,846,470 |
| Named entities | 81.88 | 4,530,368 |
| Words | 1,912.65 | 105,821,243 |
| Sentences | 85.41 | 4,726,003 |
| Part-of-speech tags and lemma | 1883.5 | 104,208,536 |
| Terms | 250.76 | 13,874,089 |

Table 3: Average and total numbers of linguistic units.

The annotation of the documents was completed in 35 hours. Table 3 shows the total number of entities found in the document collection. 106 million words and 4.72 million sentences were processed; 4.53 million named entities and 13.9 million domain specific phrases were identified. Each document contains, on average, 1,913 words, 85 sentences, 82 named entities and 251 domain specific phrases. 147 documents contained no words at all; they therefore underwent the tokenization step only. One of our NLP clients processed a 414,995 word document.

Table 4 shows the average processing time for each document. Each document has been processed in 37 seconds. Due to the exploited resource, the most time-consuming steps are the term tagging (56% of the overall processing time) and the named entity recognition (16% of the overall processing time).

|  | Average time processing | Percentage |
|---|---|---|
| loading XML input doc. | 0.38 | 1.02 |
| tokenization | 0.7 | 1.88 |
| named entity recognition | 6.12 | 16.42 |
| word segmentation | 5.19 | 13.92 |
| sentence segmentation | 0.18 | 0.48 |
| part-of-speech tagging and lemmatization | 1.84 | 4.94 |
| term tagging | 20.83 | 55.89 |
| rendering XML output doc. | 2.03 | 5.45 |
| Total | 37,27 | 100 |

Table 4: Average time for one document processing (in seconds).

The whole document collection, except two documents, has been analysed. Thanks to the distribution of the processing, the problems occuring on a specific document had no consequence on the whole process. Clients in charge of the analysis of these documents have been simply restarted.

The performance we get on this collection show the robustness of the NLP platform, and its ability to analyse large and heterogeneous collection of documents in a reasonable time. We have proven the efficiency of the overall process for semantic crawlers and its accuracy for a precise indexing of web documents.

## 6. Tuning a syntactic analyzer to the biological domain

This section presents our strategy to tune NLP tools to a given specialized domain. We take the parser as an example as its adaptation is the richest and the most complex one.

In order to extract structured pieces of information from texts, one needs to link isolated chunks of texts together. Most of the time, chunks of texts correspond to named entities and relations are expressed through verbs or predicative nouns. We thus need a reliable and precise analysis of syntactic relations between phrases. For those reasons, we chose to integrate a symbolic dependency-based parser seemed (in contrast with a constituent-based parser).

Instead of redeveloping new parsers for each sublanguage, we try to define a method for adapting a general parser to a specific sublanguage. This section presents a strategy to adapt the Link Parser (LP) (Sleator & Temerley, 1993) to parse Medline abstracts dealing with genomics. More details are given in (Aubin *et al.*, 2005).

## 6.1. The initial parser choice

LP presents several advantages among which the robustness, the good quality of the parsing, the underlying dependency formalism and the declarative format of its lexicon.

In order to test various parsers, a corpus has been built from Medline[4] abstracts (in English) dealing with transcription in *Bacillus subtilis* (Aubin *et al.*, 2005), named henceforth MED-TEST corpus. Our test corpus contains 212 randomly selected sentences (5,992 words), which contain an average of 25.4 words (from 8 to 59). Despite its relatively small size, this corpus is a good sample of the sublanguage of genomics. Medline abstracts present the following characteristics: they are made of long and syntactically complex sentences, specialized vocabulary, scientific notations and numerous non grammatical constructions.

From the results of the evaluation that we did on different parsers, it turned out that dependency-based parsers have better results on long and complex sentences, particularly with coordination. For example, LP seems to offer better performance than a constituent-based parser applied on Medline abstracts (see (Grover, 2004) for an experiment using a GPSG parser). This conclusion is shared by (Ding, 2003) who also worked on the same kind of corpus. Other experiments have also shown that the parser performance varies from one corpus to another. In the context of the ExtrAns project, (Molla et al., 2000) showed that 76% of 2,781 sentences from a Unix manpage corpus were completely parsed by LP independently to the parsing quality, while we reach only 54% on the biological corpus. When looking at the quality of the parses, we noticed different kinds of errors depending either on the biological domain or on more general linguistic difficulties like ambiguous constructions. We propose three solutions to address these issues: text normalization, terminology analysis and lexicon/grammar adaptation.

## 6.2. Diagnosis and adaptation

Our analysis of the performance of the Link grammar on the biological corpus confirms previous works. The main problems can be classified along the following axes.

*Textual noise*

Scientific texts present particularities that we chose to handle in a normalization step prior to the parsing. First, the segmentation in sentences and words was taken off from the parser and enriched with named entities recognition and rules specific to the biological domain. We also delete some extra-textual information that alters parsing quality (such as citations, for instance).

*Unknown words*

In a corpus made of full Medline abstracts, we identified 6,005 out-of-lexicon forms (45,804 occurrences) among 12,584 distinct words, *i.e.* 47.72%. They are mostly latin words, numbers, DNA sequences, gene names, misspellings and technical lexicon.

However, LP includes a module that can assign a syntactic category to an unknown word. It is based on the word suffix. Modifying the morpho-guessing (MG) module seemed a better strategy than extending the dictionary since biological objects differ from an organism to another (Grover (2004) also reports a similar process). We then created 19 new MG classes for nouns (*–ase, -ity*, etc.) and adjectives (*–al, -ous*, etc.) along with their rule. At the same time, we added about 500 words of the biological domain to the LP lexicon in different classes, mainly nouns, adjectives and verbs.

---

[4] http://www.ncbi.nlm.nih.gov/entrez/query.fcgi.

*Specific constructions*

Some words already defined in the LP lexicon present a specific usage in biological texts, which implied some modifications including moving words from one class to another and adapting or creating rules.

The main motivation for moving words from one class to another is that the abstracts are written by non-native English speakers. This point was also raised by (Pyysalo *et al.*, 2004). One way to allow the parsing of such ungrammatical sentences is to relax constraints by moving some words from the countable to the mass-countable class for instance. Some very frequent words present idiosyncratic uses (particular valency of verbs for instance), which induced the modification or creation of rules. Numbers and measure units are omnipresent in the corpus and were not necessarily well described or even present in the lexicon/grammar.

*Structural ambiguity*

We identified two cases of ambiguity that can be partially resolved by exploiting terminological information.

Prepositional attachment is a tricky point that is often fixed using statistical information from the text itself (Hindle & Rooth, 1993), a larger corpus (Bourigault & Frérot, 2004), the web (Volk, 2002) or external resources such as WordNet (Stetina & Nagao, 1997).

The second major ambiguity factor is the attachment of series of more than two nouns. like in *two-component signal transduction systems*. We noticed that such cases often appear inside larger nominal phrases often corresponding to domain specific terms. For this reason, we decided to identify terms in a pre-processing step and to reduce them to their syntactic head. If needed, the internal analysis of terms is added to the parsing result for the simplified sentence. The strategy proposed by (Sutcliffe *et al.*, 1995) that consists in the linkage of the words contained in a compound (for instance *sporulation_process*) was excluded, as it increases the lexicon size augment without reducing the parsing complexity.

Before practically integrating the use of terminology in our processing suite, we made a simulation of this simplification of terms.

## 6.3. Evaluation

We performed a two-stage evaluation of the modifications in order to measure the respective contribution of the LP adaptation on the one hand and of the term simplification on the other hand.

*Corpus and criteria*

We used a subset (10 files[5]) of the MED-TEST corpus but, contrary to the first evaluation designed for choosing a parser, we wanted to measure the quality of the whole parse and not only of specific relations.

Table 1 (for the MED-TEST subset) shows the way that out-of-lexicon words (OoL), i.e. unknown (UW) and guessed (GW) words, are handled by giving the percentage of incorrect morpho-syntactic category assignments with the original resources (lp), those adapted to biology (lp-bio) and finally the latter associated with the simplification of terms (lp-bio-t).

---

[5] 141 sentences, 2,630 words.

|     | lp  |       | Lp-bio |       | lp-bio-t |       |
|-----|-----|-------|--------|-------|----------|-------|
|     | a   | b     | a      | B     | a        | b     |
| UW  | 244 | 41.1% | 53     | 52.8% | 26       | 19.2% |
| GW  | 24  | 4.2%  | 72     | 0%    | 31       | 0%    |
| OoL | 268 | 38%   | 125    | 22.4% | 57       | 8.8%  |

Table 1: Incorrect MS category assignments
(a : total MS assignments ; b% : incorrect assignments).

In Table 2, five criteria inform on the parsing time and quality for each sentence : the number of linkages (NbL), the parsing time (PT) in seconds, the fact that a complete linkage is found or not (CLF), the number of erroneous links (EL) and the quality of the constituency parse (CQ). NbW is the average number of words in a sentence which varies with term simplification. The results are given for each one of the three versions of the parser.

| Crit. | lp      | Lp-bio  |        | lp-bio-t |        |
|-------|---------|---------|--------|----------|--------|
|       | Avg     | Avg     | %/lp   | Avg      | %/lp   |
| NbW   | 24.05   | 24.05   | 100%   | 18.9     | 78.6%  |
| NbL   | 190,306 | 232,622 | 122,2% | 1,431    | 0.75%  |
| PT    | 37.83   | 29.4    | 77.7%  | 0.53     | 1.4%   |
| CLF   | 0.54    | 0.72    | 133%   | 0.77     | 142.6% |
| EL    | 2.87    | 1.91    | 66.5%  | 1.15     | 40.1%  |
| CQ    | 0.54    | 0.7     | 129.6% | 0.8      | 148.1% |

Table 2: Parsing time and quality.

UW, GW, NbL, PT and CLF are objective data while EL and CQ necessitate linguistic expertise. The CQ evaluation consisted in the assignment of a general quality score to the sentence.

*Results and comments*

The extension of the MG module reduced the number of erroneous morpho-syntactic category assignments (see Table 1) from 38% to 22.4%. 61% of the sentences where one or more assignment error was corrected by the MG module actually have better parsing results (15% have been degraded). More generally, guessing more forms makes category assignment more reliable.

The extension of the lexicon discharged the two modules from 143 assignments out of 268 (50 of which were wrong). 64% of the sentences where one or more assignment error was corrected by the extension of lexicon have better parsing results (18% of the sentences were degraded).

The effect of rule modification and creation is difficult to evaluate precisely though it actually improves the parsing, especially by relaxing the constraints on determiners and inserts.

The most obvious contribution to the better parsing quality is the one of term simplification. The drastic reduction in parsing time and number of linkages gives an idea of the reduction of complexity. It is not only due to the smaller number of words since the number of erroneous links is reduced by 60% while the number of words is reduced by only 21.4%. This confirms previous similar studies that showed a reduction of 40% of the error rate on the main syntactic relations with a French corpus.

The remaining errors are due to four different phenomena. First, the normalization step, prior to parsing, needs to be enhanced. Concerning LP, there are still lexicon gaps, wrong class assignments and a still unsatisfactory handling of numerical expressions. In addition, and like (Sutcliffe *et al.*, 1995), we identified a weakness of LP regarding coordination. A specific study of the coordination system in LP and in the biological texts may be necessary. Finally, some ambiguous nominal and prepositional attachments still remain in spite of term simplification. These may be resolved in a post-processing step like in ExtrAns that uses a corpus based approach to retrieve the correct attachment from the different linkages given by LP for a sentence.

Thus, the parser adaptation relies on three methods: the exploitation of a small base of morphological rules, the modification of the grammar, and an adequate integration that relieve the parser from all what do not directly deal with structural ambiguity (POS and term tagging, especially).

## 7. Conclusion

We have presented in this paper a platform that has been designed to enrich specialized domain documents with linguistic annotations. While developments and experiments have been performed on biomedical texts, we assume that this architecture is generic enough to process other specialized documents. The platform is designed as a framework using existing NLP tools which can be substituted by others if necessary. Several NLP modules have been integrated: named entity tagging, word and sentence segmentation, POS tagging, lemmatization, term tagging, and syntactic parsing. Semantic type tagging and anaphora resolution are currently being under stress.

We also focused on the system performance, since this point is crucial for most Internet applications. We have experimented a distributed design of the platform, by splitting the corpus in equal parts: this strategy dramatically increased the overall performance (see (Ravichandran *et al.*, 2004). We have also shown that Ogmios is a robust NLP platform with respect to the high heterogeneity of the document sizes and types.

These first experiments show that a deep analysis of web documents is possible. Besides the necessary improvement the Ogmios platform, our next goal is to assess the impact of NLP on IR performance. Our hypothesis is that this impact should be higher in the case of a specialized search engines than for a generic IR framework, on which the IR-NLP cooperation has mainly been tested until now. Specific experiments are currently carried out in the ALVIS project to test the potential resulting enhanced functionalities on a microbiological search engine.

**Acknowledgements**

This work is supported by the EU 6th Framework Program in the IST Priority under the ALVIS project. The material benefits from interactions with the ALVIS partners, especially with INRA-MIG.